# A Standalone Markerless 3D Tracker for Handheld Augmented Reality

João Paulo Lima • Veronica Teichrieb • Judith Kelner

*Virtual Reality and Multimedia Research Group, Informatics Center, Federal University of Pernambuco*

*Av. Prof. Moraes Rego S/N, Prédio da Positiva, 1° Andar, Cidade Universitária, 50670-901, Recife – PE – Brazil*

+55 81 2126-8954

+55 81 2126-8955

{jpsml, vt, jk}@cin.ufpe.br

**Abstract** This paper presents an implementation of a markerless tracking technique targeted to the Windows Mobile Pocket PC platform. The primary aim of this work is to allow the development of standalone augmented reality applications for handheld devices based on natural feature tracking. In order to achieve this goal, a subset of two computer vision libraries was ported to the Pocket PC platform. They were also adapted to use fixed point math, with the purpose of improving the overall performance of the routines. The port of these libraries opens up the possibility of having other computer vision tasks being executed on mobile platforms. A model based tracking approach that relies on edge information was adopted. Since it does not require a high processing power, it is suitable for constrained devices such as handhelds. The OpenGL ES graphics library was used to perform computer vision tasks, taking advantage of existing graphics hardware acceleration. An augmented reality application was created using the implemented technique and evaluations were done regarding tracking performance and accuracy.

**Keywords** Augmented reality • Computer vision • Markerless tracking • Handheld

## 1 Introduction

Two topics have been gaining more attention of Augmented Reality (AR) researchers in the latest years: handheld devices support and markerless tracking. Mobility and compactness requirements of some application domains favor AR projects that focus on handheld platforms. In addition, markers placed all around in the environment can be considered intrusive, justifying the need for natural feature tracking.

Handheld based AR systems can be classified as distributed or standalone. In distributed systems, part of the tasks needed to augment the environment is performed by a server, which exchanges data with the mobile device. There can be different levels of distribution, as explained in [18]. On the other hand, standalone systems do not rely on a server for generating the AR output, implementing all the required procedures by themselves.

Existing full-featured markerless AR solutions for handheld platforms are distributed and the tracking phase is performed on the server side. The server usually has a great processing speed and additionally, it can also have a powerful Graphics Processing Unit (GPU). However, each frame captured by the mobile device is transferred to the server in order to be processed. The results are then sent back to the handheld, which may cause a delay that impacts the frame rate of the applications. Another issue is related to the need for a computer server, which harms the mobility of the solutions.

This paper presents a standalone markerless AR solution running on the Microsoft Windows Mobile Pocket PC handheld platform which promotes the development of fully mobile AR applications. A modified version of an edge based markerless tracking algorithm, which is described in [21], was adapted to run under the constrained processing and graphics capabilities of a handheld device. As a byproduct, a portion of the Vision-something-Libraries (VXL) [16] and the Visual Servoing Platform (ViSP) [8] was ported to the Pocket PC platform, providing an infra-structure for the development of other computer vision solutions targeted at mobile devices. A method for the edge visibility checking step, which makes use of OpenGL ES, was also developed. It is suitable for the reduced functionality of the given mobile graphics library. It can also exploit the graphics hardware acceleration available on some handheld devices.

Most AR tracking systems working on mobile devices are based on the use of fiducial markers [5, 17]. When markers are not an option due to application requirements, a tracking approach based on natural features should be considered. Techniques developed for online monocular markerless AR can be classified into two major types: model based and Structure from Motion (SfM) based. Model based techniques require that knowledge about the real world is obtained before tracking occurs. This knowledge is stored in a 3D model used for estimating camera pose. In SfM based approaches, camera movement throughout the frames is estimated without any previous knowledge about the scene. The knowledge is acquired during tracking. Model based techniques can be classified into three categories: edge based, where camera pose is estimated by matching a wireframe 3D model of an object with the real world image edge information [21]; optical flow based, which exploits temporal information extracted from the relative movement of the object projection onto the image in order to track it [1]; and texture based, which takes into account texture information available in images for tracking [15]. In this paper a model based tracking method that belongs to the edge based category has been implemented.

There were some attempts to perform markerless tracking on mobile devices. Mozzies [14] is a First Person Shooting (FPS) game where camera movement is detected based on the captured image in order to allow the player to aim at the enemies. Kick Real [11] consists in a penalty shootout competition where the player kicks the virtual ball with his own foot. However, the previously mentioned games only perform 2D tracking of the environment. Virtual Video [6] uses a cell phone equipped with a Global Positioning System (GPS), accelerometers and a magnetometer to determine the position and orientation of the camera in relation to the real world. Nevertheless, the tracking is not very accurate and provides only an approximation of the objects' location. AR-PDA [3] and AR Phone [20] are able to perform precise 3D natural feature tracking. However, they are distributed, since all computer vision processing is done by a server. The ULTRA system [12] performs autonomous 3D markerless calibration, although not in real-time. A standalone natural feature tracking solution for mobile phones that runs at interactive frame rates is described in [19], but it is only able to detect planar objects. The solution proposed in this work is capable of performing 3D markerless tracking in a standalone manner, using a handheld device.

This paper is organized as follows. Section 2 details the edge based tracking algorithm pipeline used as the basis for this work. Section 3 explains the steps needed to implement the markerless tracking feature on the Pocket PC platform. The results obtained and evaluations performed are presented in Section 4. Section 5 draws some final considerations and discusses possible future work.

## 2 Edge based tracking

The work described in this paper implements a variation of the single hypothesis edge based tracking technique by Wuest et al. [21]. This technique was chosen because it utilizes consolidated methods in the markerless tracking area and also due to its low CPU load, being able to run on mobile devices in an autonomous way.

The first step of the edge based tracking is to determine what are the visible parts of the edges when the 3D model of the object is projected on the image plane using the previous pose information. The visibility checking applied by Wuest et al. could not be used on the handheld platform and a different approach was developed, as further explained in Subsection 3.3.

A sampling of the projected visible edges is done, obtaining control points, as can be seen in Figure 1. This is performed in image space in order to get evenly spaced points. The number of sampled points per edge n is calculated using the following formula:

$$n = \frac{\text{edge length}}{\text{sampling step}}. \tag{1}$$

The sampling step value used in the current implementation was empirically set to 10.

Next, for each control point, it is determined a correspondent point in the image gradient. This search is done in a perpendicular direction in relation to the edge, as shown in Figure 1, where the green lines represent strong gradients in the image. Differently from Wuest et al., which uses a simple filter mask, the Moving Edges (ME) algorithm [2] was utilized for finding the correspondences. Since the ME algorithm works with gray scale images, the input colored image has to be previously converted.

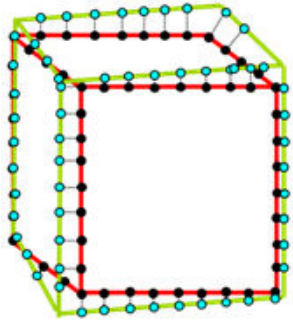

**Fig. 1** Control points sampling and correspondences search

Since the correspondences between points and edges are known, the Levenberg-Marquardt method is used to calculate the pose minimizing the reprojection error, defined as:

$$err = \sum_i \Delta(p_i, q_i), \tag{2}$$

where $\Delta$ is the distance between the projected control point $p_i$ and the line that passes by the correspondent point $q_i$ in the image. $\Delta$ consists in:

$$\Delta(p_i, q_i) = |(q_i - p_i) \cdot (n_i)|, \qquad (3)$$

where $n_i$ is the normal of the projected edge. The pose is parameterized using six variables: three for rotation and three for translation. Instead of considering all the nine elements of a $3x3$ matrix, the rotation is represented using an exponential map. This reduces the number of variables to be minimized and avoids the introduction of unnecessary constraints to ensure that the rotation matrix is orthonormal. In the exponential map formulation, the three parameters define a vector, which represents the rotation axis. The vector norm is the rotation angle around the axis.

## 3 Handheld implementation

In order to run the edge based tracking solution on a Pocket PC, computer vision libraries that implement some required features needed to be available on the platform. Therefore, part of the VXL and ViSP libraries were ported to Pocket PC. VXL provided all math support required, including the Levenberg-Marquardt method. ViSP contains an implementation of the ME algorithm. After that, most of the math code was modified to use fixed point, aiming for performance improvements. Finally, edge visibility was handled using the OpenGL ES graphics library. The next subsections describe how these steps were accomplished.

### 3.1 VXL port for Pocket PC

VXL is an open source C++ portable project maintained by a team consisting of a number of universities (Brown University, University of Oxford, Leuven University, etc.) and companies (General Electric, Kitware, among others). It is composed of several libraries, and some of them were ported to the handheld platform.

The VXL libraries that had to be ported to Pocket PC in order to run the edge based tracking properly were: Vision C++ Compatibility Library (vcl), which acts as an abstraction layer to the main C++ features in order to ensure compatibility with existing compilers; Vision Numerics Library (vnl), for numerical algorithms and data structures; and Vision Third Party Netlib (v3p_netlib), used by vnl to access the Fortran based Netlib math software [9]. The unit tests related to these libraries, together with testlib, used for unit testing of the entire framework, were also ported to ensure that the modified code satisfies all tests on the new platform. The numerics library is split into vnl and vnl_algo, which enclose matrix and polynomial data structures and numerical algorithms, respectively. Summarizing, the full vcl, vnl and testlib projects, as well as parts of vnl_algo and v3p_netlib related to the Levenberg-Marquardt algorithm, were made available on the handheld platform.

Modifications in vcl were made due to the absence of the standard C functions abort and time on the Pocket PC platform. Therefore, all calls to abort were disabled when compiling for Windows Mobile. Concerning the time function, an implementation available on libCE [7] was used.

Regarding vnl, floating point rounding could not be done using _asm and the fistp instruction, since inline assembly is not supported on the mobile platform. Besides, the ARM processor used does not have a Floating Point Unit (FPU). Thus, the assembly routine was replaced by a trivial C code that performs the same functionality.

Complex numbers support in v3p_netlib was disabled in order to avoid compile errors on Pocket PC. This solution satisfies the case study performed in this work (edge based tracking), since the Levenberg-Marquardt algorithm implementation does not need complex numbers.

Once the required portion of the v3p_netlib library was successfully ported, vnl_algo also worked properly on the mobile platform without any changes.

Finally, two modifications were implemented in testlib in order to make it compatible with Pocket PC. The first one is related to string format conversion from wide char (Unicode) to multi byte (ASCII), because some testlib functions used ASCII strings, while Unicode is the default encoding of the mobile platform. The second modification consists in disabling the redirection of all debug information to log files. This redirection relies on the crtdbg library, which is not available for Pocket PC.

### 3.2 ViSP port for Pocket PC

ViSP is a visual servoing library created by the INRIA Lagadic research group, which is also open source and based on the C++ programming language.

The tracking, transformation estimation and image classes of ViSP related to ME were ported to the Pocket PC platform. In addition, all code associated with display functions was removed from the mobile version of the library, as they were not necessary for this work. Math constants for $\pi$, $\pi/2$ and $\pi/4$ also had to be defined, since they do not exist on the Microsoft math library. Finally, it was added a procedure to the image conversion class for gray scaling RGB565 images, as this is the standard pixel format for handheld devices.

### 3.3 Fixed point math

Since the targeted processor does not have a FPU, all floating point operations are emulated by software, which causes a performance penalty. Therefore, instead of using floating point types to perform real number calculations, a fixed point type had to be utilized. In order to accomplish this, a type for real numbers was created. Depending on the platform being used, it is mapped to fixed or floating point type. In addition, the code of the supporting libraries and the application was modified to use the real type instead of directly the floating point types.

The fixed point type provides implementation for basic operations, comparisons, absolute value, square root, rounding and trigonometric functions. The 64-bit fixed point math format initially used was S47.16 (1 bit for sign, 47 bits for the integer part and 16 bits for the decimal part). A large amount of bits was used to represent the integer part in order to avoid overflows.

One of the main drawbacks of using a fixed point representation is the lack of precision in the decimal part. As a consequence, the fixed point version of the Levenberg-Marquardt algorithm needed more iterations to converge than the floating point counterpart, resulting in a performance decrease. In addition, the quality of the minimization was also harmed. Therefore, the number of bits used to represent the decimal part was raised and the S40.23 format was adopted. As a result, it did not present overflows and the numeric optimization routines' outputs were satisfactory.

### 3.4 Visible edges detection

The approach adopted in [21] for determining the visible parts of the edges at a given frame makes use of the OpenGL extension GL_OCCLUSION_TEST_HP. However, neither this extension nor the equivalent standard extension GL_ARB_OCCLUSION_QUERY are available for OpenGL ES. In addition, reading from the depth buffer is not allowed by the mobile graphics library.

An alternative method was developed to perform visibility testing on the handheld platform. It was inspired by the Facet-ID method described in [15], and as its goal is to identify edges, the technique was named Edge-ID. In Facet-ID, the index of each polygon is encoded in its color value, and after the model is rendered, it is possible to discover the facet that generated a given pixel when projected. Edge-ID exploits the same idea for edges, but for a different purpose: while Facet-ID is used for finding the 3D backprojection of a pixel and its normal at the model, Edge-ID aims to determine if a control point sampled from an edge is visible or not. Another difference between the methods is that in Facet-ID the model is drawn with filled faces, while in Edge-ID a wireframe model with hidden line removal is rendered. This way, only the visible model edges will have a color value different from the background color. It is then possible to find out if a control point $p(x, y)$ is visible by comparing the index of its edge with the index decoded from the color stored at the position $(x, y)$ in the color buffer. The use of unique IDs for each edge is justified by the fact that points from different edges can be projected at the same position in image space. If no ID checking is performed, a hidden control point could be considered visible. Figure 2 illustrates the proposed visibility testing approach.

In summary, the outline of the Edge-ID method is as follows:

1 Map the color value of each model edge to its index.

2 Render the model edges with hidden line removal.

3 For each model edge $i$

    3.1 Sample the edge, obtaining control points

    3.2 For each sampled point $p(x, y)$

        3.2.1 If $ID(x, y) = i$, then the point is visible

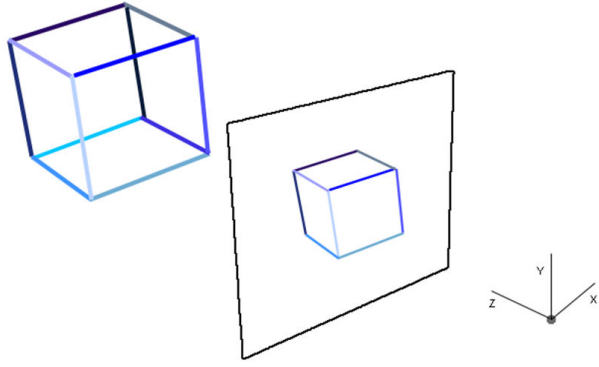

**Fig. 2** Edge-ID method

Initially, the coding scheme adopted for mapping the IDs to RGB color components was rather simple. Black color ($R = 0$, $G = 0$, $B = 0$) is reserved for representing the background. Then, each edge index is incremented by one and, considering its 32-bit binary representation, the most significant byte is stored at the red channel, the next byte is stored at the green channel and the least significant byte is stored at the blue channel. The inverse process is done for decoding. With this representation, the maximum number of model edges is $2^{32} - 1 = 4,294,967,295$. The average edge count of the models commonly used for tracking does not get even close to this value.

Using this coding scheme on the handheld platform presented some problems related with OpenGL ES. In the available graphics library implementation, the primitives are not rendered with the exact color value specified for it. Instead, an approximation is done and a color value close to the original one is used. This leads to confusion between different edges that are drawn with the same color. The solution found to this problem was to choose uniformly spaced values in color domain for representing the edges. A spacing of 8 levels between consecutive component values has shown to be sufficient in order to prevent confusion between colors. This way, an edge index $i$ is encoded using the following equations:

$$b_{code} = (i + 1) \cdot 8, \tag{4}$$

$$g_{code} = (b_{code} / 256) \cdot 8, \tag{5}$$

$$r_{code} = (g_{code} / 256) \cdot 8, \tag{6}$$

$$R = r_{code} \bmod 256, \tag{7}$$

$$G = g_{code} \bmod 256, \tag{8}$$

$$B = b_{code} \bmod 256. \tag{9}$$

The edge index can then be decoded by:

$$rg_{decode} = (R/8) \cdot 256 + G, \tag{10}$$

$$rgb_{decode} = (rg_{decode}/8) \cdot 256 + B, \tag{11}$$

$$i = (rgb_{decode}/8) - 1. \tag{12}$$

This coding scheme is capable of representing at most $2^{15} - 1 = 32{,}767$ edges, which is still sufficient for 3D tracking applications.

## 4 Results

A simple standalone application that tracks a 3D cube object and displays a solid cone model registered with it was developed for the handheld platform. It applied the described edge based tracking algorithm and the computer vision infrastructure. The mobile device used in the tests was a Personal Digital Assistant (PDA) Dell Axim x51v. It has a 624 MHz Intel PXA270 XScale ARMV5 processor, 256 MB of ROM, 64 MB of RAM, and a VGA LCD display with 16 bit color depth. This PDA also has an Intel 2700G multimedia accelerator with 16 MB of video memory. The operating system is Microsoft Windows Mobile 5.0 Pocket PC. The camera used on the mobile device was the Spectec SD (Secure Digital) Camera SDC-001A, with QVGA resolution and a frame rate of 15 fps (frames per second). The development tool employed to implement the project was Microsoft Visual Studio .NET 2005 Professional Edition. Hardware accelerated OpenGL ES version 1.0 was utilized for 3D graphics rendering, and the GLUT|ES library [4] was in charge of GUI, application loop and input handling.

Once working, the application was evaluated taking into account frame rate, accuracy and robustness metrics. Initially, synthetic QVGA images were used as input, which were generated with the help of Object-oriented Graphics Rendering Engine (OGRE) [10]. Figure 3 shows some pose estimation results obtained on the handheld platform.

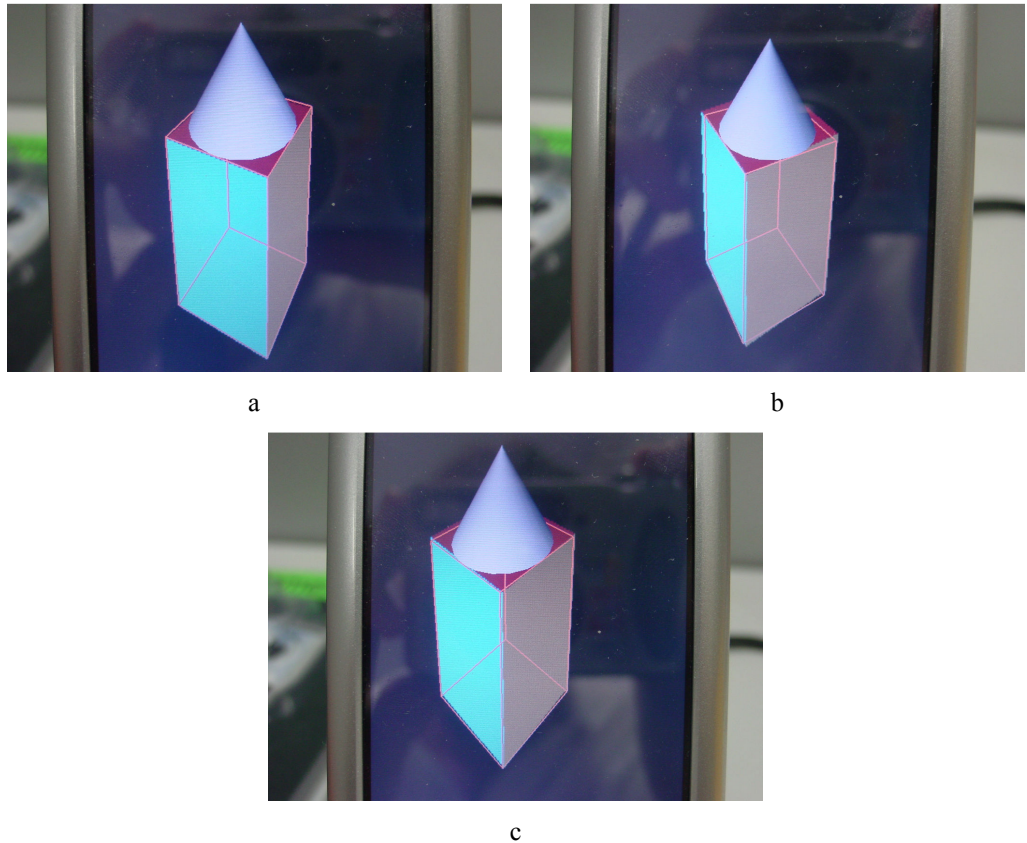

**Fig. 3** Tracking results for frames 0 (a), 35 (b), and 45 (c) of the synthetic sequence

Table 1 presents the percentage of time required by each step of the tracking algorithm running on the handheld device and using the cube sequence mentioned above as input. The average total time spent for tracking a frame is 64 ms, which results in a 15.625 fps rate. Around 60 points are tracked during the sequence. Figure 4 shows the total times spent for tracking each of the first 60 synthetic frames. The obtained frame rate is adequate to AR applications, especially the handheld targeted ones.

**Table 1** Percentages of computation time for each step of the tracking algorithm on the handheld platform

| STEP | TIME (%) |
|---|---|
| **Visible edges detection** | **28** |
| **Image gray scaling** | **19** |
| **ME** | **37** |
| **Pose calculation** | **16** |

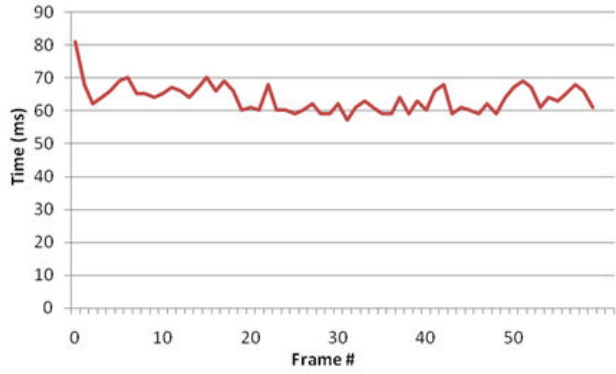

**Fig. 4** Total computation times for each of the first 60 frames of the synthetic sequence

The camera position calculated by the tracking algorithm in the $x$, $y$ and $z$ axis are presented in Figure 5, together with the corresponding ground truth values.

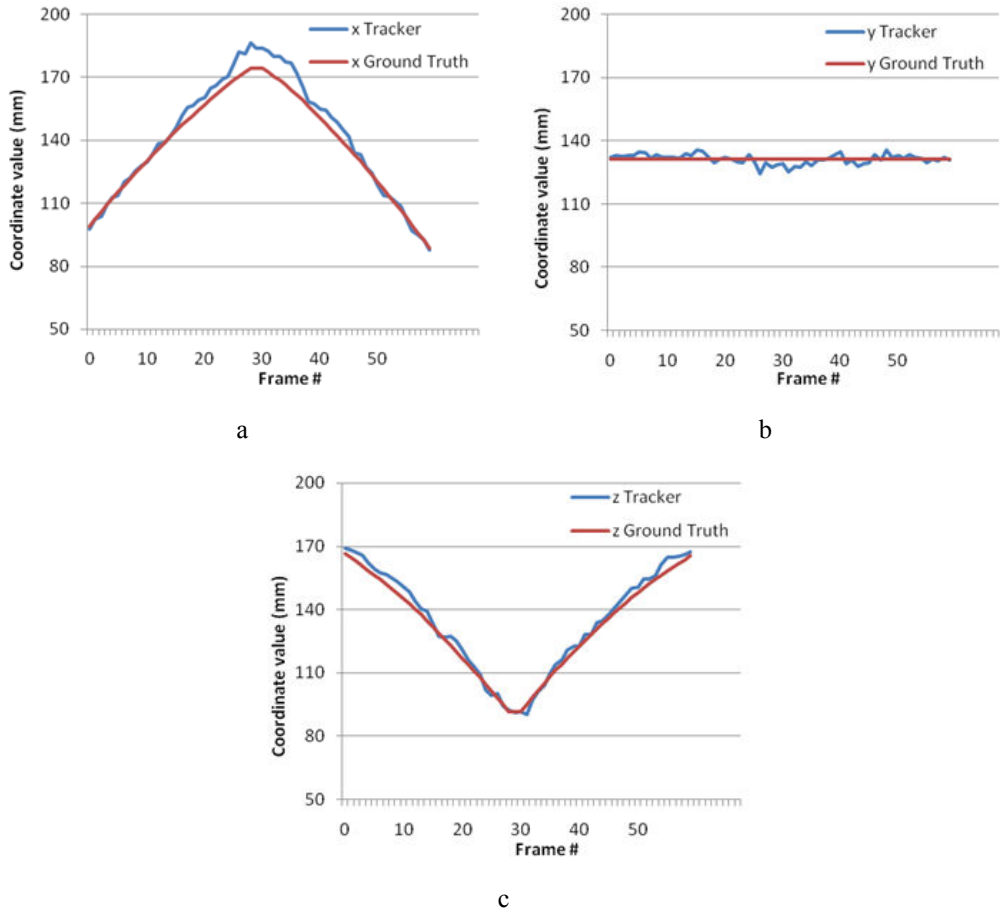

**Fig. 5** Estimation accuracy of camera position in the $x$ (a), $y$ (b) and $z$ (c) axis

The average errors were the following: 4.10 mm in the $x$ axis, 1.79 mm in the $y$ axis and 2.73 mm in the $z$ axis. The average distance between the calculated camera position and the ground truth was 6.12 mm. The distance between the tracked object and the camera was about 150 mm. The side length of the cube was 60 mm. Since visual perception is very important in AR applications, it is reasonable to say that the obtained error rates are acceptable. While the pose estimation error is not very visually perceptible, tracking accuracy should still be improved.

After the first tests with synthetic data, the handheld edge based tracker was evaluated using images of the real world captured by a camera. Figure 6 depicts some augmentation results. The tracker showed to be robust to a certain level of occlusion of the tracked object. The cube model has 12 contour edges and was tracked at 15 fps. The wood toy model has 30 contour edges and was tracked at 10 fps.

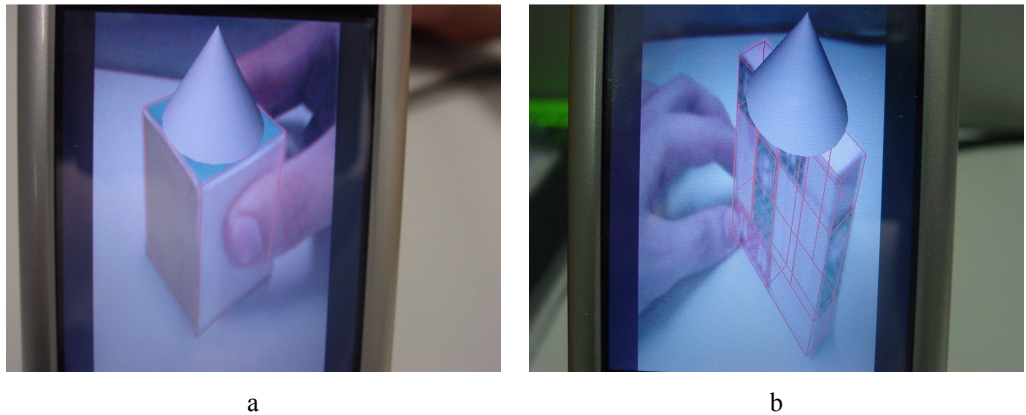

a　　　　　　　　　　　　　　　b

**Fig. 6** Augmented cube (a) and wood toy (b) on the handheld platform

## 5 Conclusions and future work

An edge based tracking solution for the Pocket PC platform has been developed, which makes handheld based standalone markerless AR applications feasible. This was accomplished building a computer vision infra-structure for the Pocket PC platform through a partial port of the VXL and ViSP libraries. Fixed point math was also applied to most calculations, focusing better performance. These outcomes are byproducts of this work and can be applied to develop other computer vision solutions targeting mobile devices.

The frame rate obtained with the test application is suitable for handheld AR. In addition, the system provides a reasonable estimation of the object pose, visually speaking. However, tracking accuracy can be improved.

As future work, robust estimators could be used to improve tracking robustness and accuracy. Multiple correspondence hypotheses might be considered in the ME algorithm, aiming at a more robust pose estimation procedure. The implementation of an automatic tracker initialization method such as [13] is also planned for the handheld platform, since it is currently being done manually.